\documentclass[11pt,a4paper]{article}
\usepackage[hyperref]{ranlp2021}
\usepackage{times}
\usepackage{latexsym}
\usepackage{graphicx} 
\usepackage{amsmath}
\usepackage{todonotes}

\usepackage{xcolor}
\usepackage{soul}
\usepackage[T1]{fontenc}

\usepackage{microtype}

\aclfinalcopy %
\def\aclpaperid{***} %

\usepackage{soul}

\newcommand{\cut}[1]{} %

\newenvironment{compact_itemize}
{\begin{itemize}\vspace*{-0.2cm}\setlength{\itemsep}{-2pt}}
{\vspace*{-0.2cm}\end{itemize}}

\usepackage{multirow}
\usepackage{booktabs} %

\title{Not All Comments are Equal: \\ Insights into Comment Moderation from a Topic-Aware Model}

\author{Elaine Zosa \\
  University of Helsinki \\
  \texttt{elaine.zosa@helsinki.fi}\\ 
  \And Ravi Shekhar \\
  Queen Mary University of London \\
  \texttt{r.shekhar@qmul.ac.uk} \\ 
  \AND 
    Mladen Karan \\
  $^{\diamondsuit}$Queen Mary University of London \\
  \texttt{m.karan@qmul.ac.uk} \\ \And
      Matthew Purver$^{\diamondsuit,\dagger}$ \\
  $^{\dagger}$Jožef Stefan Institute \\
  \texttt{m.purver@qmul.ac.uk} \\}

\date{}

\begin{document}
\maketitle
\begin{abstract}
 
Moderation of reader comments is a significant problem for online news platforms. Here, we experiment with models for automatic moderation, using a dataset of comments from a popular Croatian newspaper. Our analysis shows that while comments that violate the moderation rules mostly share common linguistic and thematic features, their content varies across the different sections of the newspaper. We therefore make our models topic-aware, incorporating semantic features from a topic model into the classification decision. Our results show that topic information improves the performance of the model, increases its confidence in correct outputs, and helps us understand the model's outputs. 
\end{abstract}

\section{Introduction}
\label{sec:intro}

Most newspapers publish their articles online, and allow readers to comment on those articles. This can increase user engagement and page views, and provides readers with an important route to public freedom of expression and opinion, with the ability to interact and discuss with others. Comment sections usually provide some degree of anonymity;\footnote{Some newspapers allow completely anonymous posting; some require commenters to create an account with a username, but this does not usually reveal their true identity.} while improving accessibility, this can also encourage inappropriate behaviour, and publishers therefore usually employ some moderation policy to regulate content and to ensure legal compliance (in some cases, publishers can be held responsible for user-contributed content on their sites).

One possible approach is a `moderate then publish' policy, in which comments must be approved by a moderator before they appear; this requires significant manpower and introduces delays and limitations into the user conversation (for example, the New York Times %
only allows comments for one day after article publication\footnote{NYT Comment FAQ: https://nyti.ms/2PF02kj}). On the other hand, a `publish then moderate' strategy, in which comments are published immediately, and later removed if necessary, is less effective at blocking toxic or illegal content.
Combined with the increase in comment volumes in recent years %
there is increasing interest in automatic moderation methods \citep[see e.g.][]{pavlopoulos-etal-2017-deeper}, either as stand-alone tools or for integration into human moderators' practices \citep{schabus-skowron-2018-academic}. 

Detecting comments that need moderators' attention is usually approached as a text classification task \citep[see e.g.][]{pavlopoulos-etal-2017-deeper}; but comments can be blocked for a range of reasons \citep{shekhar-etal-2020-automating}. One is the presence of offensive language, a well-studied NLP task (see Section~\ref{sec:relWork} below); %
however, others include advertising or spam, illegal content, spreading misinformation, trolling and incitement --- all distinct categories which might be expected to show distinct features, and perhaps to vary according to the content being commented on.
Another aspect that distinguishes the comment moderation task from the usual text classification tasks in NLP is the need for interpretable or explainable models: if classifiers are to be used by human moderators within publishers' working practices, they must be able to understand the outputs \citep[][]{svec-etal-2018-improving}. 

Here, we therefore investigate models which can provide both an aspect of interpretability and the ability to take account of the topics being discussed, by incorporating topic information into the comment classifier. Specifically, we incorporate semantic representations learned by the Embedded Topic Model (ETM)~\citep{dieng-etal-2020-topic} into a classifier pipeline based on Long Short-Term Memory (LSTM) networks~\citep{hochreiter1997long}. Our model improves performance by 4.4\% over a text-only approach on the same dataset~\citep{shekhar-etal-2020-automating}, and is more confident in the correct decisions it makes. Inspection of the topic distributions reveals how different newspaper sections have different language and topic distributions, including differences in the kind of comments that need moderation.\footnote{Source code available at \url{https://github.com/ezosa/topic-aware-moderation}} %

\section{Related Work}
\label{sec:relWork}

\paragraph{Automated news comment moderation}

Most research on this task so far formulates it as a text classification problem: for a given comment, the model must predict whether the comment violates the newspaper's policy. However, approaches to classification vary. \citet{Nobata2016AbusiveLD} use a range of linguistic features, e.g.\ lexicon and n-grams. \citet{pavlopoulos-etal-2017-deeper} and \citet{svec-etal-2018-improving} use neural networks, specifically RNNs with an attention mechanism.  Recently, \citet{tan-etal-2020-tnt} and \citet{tran-etal-2020-habertor} apply a modified BERT model~\citep{devlin-etal-2019-bert} while \citet{schabus2017one} use a bag-of-words approach.

Some approaches go beyond the comment text itself: \citet{gao-huang-2017-detecting} add information like user ID and article headline into their RNN to make the model context-aware; \citet{pavlopoulos-etal-2017-improved} incorporate user embeddings; \citet{schabus-skowron-2018-academic} incorporate the  news category metadata of the article. However, no work so far investigates automatic modelling of topics (rather than relying on categorical metadata), or applies this to the comments rather than just their parent articles.

Some steps towards model intepretability and output explanation have also been taken: both \citet{svec-etal-2018-improving} and \citet{pavlopoulos-etal-2017-deeper} use an attention saliency map to highlight possibly problematic words. However, we are not aware of any work using higher-level topic information as a route to understanding model outputs.

\paragraph{Available datasets} Several datasets have been created for the news comment moderation task. \citet{Nobata2016AbusiveLD} provide 1.43M comments posted on Yahoo! Finance and News over 1.5 years, in which 7\% of the comments are labelled as abusive via a community moderation process. \citet{gao-huang-2017-detecting} contains 1.5k comments from Fox News, annotated with specific hateful/non-hateful labels as a post-hoc task, and having 28\% hateful comments. However, both are relatively small, and their labelling methods mean that neither dataset is entirely representative of the moderation process performed by newspapers. 

\citet{pavlopoulos-etal-2017-deeper} provides 1.6M comments from Gazzetta, a Greek sports news portal, over c.1.5 years. Here, 34\% of comments are labelled as blocked, and the labels are derived from the newspaper’s human moderators and journalists. \citet{schabus2017one} and  \citet{schabus-skowron-2018-academic} provide a dataset from a German-language Austrian newspaper with 1M comments posted over 1 year, out of which 11,773 comments are annotated using seven different rules. %

More recently, \citet{shekhar-etal-2020-automating} present a dataset from 24sata, Croatia's most widely read newspaper.\footnote{\url{http://24sata.hr/}} This dataset is significantly larger (10 years, c.20M comments); and moderator labels include not only a label for blocked comments, but also a record of the reason for the decision according to a 9-class moderation policy. However, their experiments show that classifier performance is limited, and transfers poorly across years. Here, we therefore use this dataset (see Section~\ref{sec:dataset}), with a view to improving performance and applying a topic-aware model to  improve and better understand the robustness in the face of changing topics.

\paragraph{Related tasks}

More attention has been given to related tasks, most prominently the detection of offensive language, hate speech, and toxicity~\citep{pelicon2021investigating}. A comprehensive survey of dataset collection is provided by~\citet{poletto2020resources} and~\citet{vidgen2020directions}.\footnote{\url{http://hatespeechdata.com/} provides a comprehensive list of relevant datasets.} %

\paragraph{Topic Modelling}

Topic models capture the latent themes (also known as \textit{topics}) from a collection of documents through the co-occurence statistics of the words used in a document. Latent Dirichlet Allocation (LDA) \cite{blei2003latent}, a popular method for capturing these topics, is a generative document model where a document is a mixture of topics expressed as a probability distribution over the topics and a topic is a distribution over the words in a vocabulary. The Embedded Topic Model \citep[ETM,][]{dieng-etal-2020-topic} is an LDA-like topic modelling method that exploits the semantic information captured in word embeddings during topic inference. The advantage of ETM over LDA is that it combines the advantages of word embeddings with the document-level dependencies captured by topic modelling and has been shown to produce more coherent topics than regular LDA.

\section{Dataset}
\label{sec:dataset}

We use the 24sata comment dataset~\citep{shekhar-etal-2020-automating,pollak-etal-2021-embeddia}, %
introduced in Section~\ref{sec:relWork}. 
This contains c.21M comments on 476K articles from the years 2007-2019\footnote{Dataset is available at http://hdl.handle.net/11356/1399}, written in Croatian. The dataset has details of comments blocked by the 24sata moderators, based on a set of moderation rules--these vary from hate speech to abuse to spam \citep[see][for rule description]{shekhar-etal-2020-automating}. The dataset also identifies the article under which a comment was posted, together with the section/sub-section of the newspaper the article appeared in. These sections/sub-sections relate to the content of the article: for example, the Sport section contains sports-related news while the Kolumne (\textit{Columns}) section contains opinion pieces. The largest section, Vijesti (\textit{News}), is further subdivided as shown in Table \ref{tab:dataset-2019}.

\begin{table}
\begin{tabular}{| l | r | r | r |}
\hline
\multicolumn{4}{|c|}{Comment Moderation Data}
\\\hline
{} & Blocked & Non-blocked & Blocking Rate\\

\hline
Train   & 4984 & 75016  & 6.23\% \\
Valid 	& 642 & 9358 &  6.42\% \\
Test	& 37271 & 438142  &  7.84\%\\
\hline
\hline
\end{tabular}
\\
\begin{tabular}{| l | r | r | r|}
\multicolumn{4}{|c|}{Topic Modelling Data}
\\\hline

{} & Blocked & Non-blocked & Blocking Rate\\
\hline
Train & 34863 & 36725  & 48.70\% \\
Valid &	4880  & 5120 & 48.80\% \\
\hline
\end{tabular}

\caption{Details of datasets used in experiments.}
\label{tab:dataset-train}
\end{table}

\subsection{Data Selection}
\label{sec:data-selection}

In this work, we use data from 2018 for training and validation of the topic model and classifiers and data from 2019 for testing. This reflects the realistic scenario where we use data collected from the past to make predictions. For training and validation, we randomly select 50,000 articles out of 65,989 articles from 2018, sampling from the nine most-representative sections/sub-sections (Table~\ref{tab:dataset-2019}). Each article comes with c.50 comments on average. 

To train the topic model, we sample around 80,000 comments across these articles, with a roughly equal split between blocked and non-blocked comments. This is to encourage a diverse mix of topics from both comment classes. As a preprocessing step we remove comments with less than 10 words from the training data (see Table~\ref{tab:dataset-train} (lower part)). To train the classifiers, we randomly sample around 80,000 comments such that the sampled set has the same blocking rate as the entire 2018 dataset.

For the test set, we then use all 475,413 comments associated with the 17,953 articles from 2019. Table~\ref{tab:dataset-train} (upper part) provides the dataset details, with comment moderation blocking rate. For the test set, Table~\ref{tab:dataset-2019} provides details on the section and sub-section of the related articles. These top nine sections account for more than 95\% of the comments of the entire test set.

\begin{table}
\resizebox{0.5\textwidth}{!}{
\begin{tabular}{l  r  r  r}
\toprule 

Section & \hspace*{-1cm}Blocked & Non- & Blocking\\ 
($~~-$ Subsection) &  & blocked &  Rate\\ 
 \hline
Kolumne (\textit{Columns})	& 655	& 6382 & 9.31\% \\
Lifestyle & 2426	& 30985  & 7.26\% \\
Show	    & 6827  & 58896  & 10.39\% \\
Sport     & 5882  & 80820  & 6.78\%  \\
Tech       & 382	& 7173  & 5.06\% \\
 \hline
Vijesti (\textit{News})   & 20094 & 239835 &7.73\%\\
\hline
$~~-$ Crna kronika (\textit{Crime}) & 5917	& 62471 & 8.65\% \\
$~~-$ Hrvatska (\textit{Croatia})     & 3527  & 45170 & 7.70\% \\
$~~-$ Politika (\textit{Politics})      & 6088	& 80264 & 7.05\% \\
$~~-$ Svijet (\textit{World})    & 2625	& 31459 & 7.24\% \\
\toprule
\end{tabular}

}
\caption{Details per section, and (for section Vijesti) sub-section, of the comment moderation test set.}
\label{tab:dataset-2019}
\end{table}

\subsection{Content Analysis}
\label{sec:analysis-content}
To gain some insight into the content of blocked comments, we analyze the linguistic differences between blocked and non-blocked comments and across different sections. First, we compare comment length. As we can see from Table~\ref{tab:msttr}, blocked and non-blocked comments have, on average, similar lengths. However, if we further divide blocked comments into two sub-groups --- spam and non-spam --- \cut{For this analysis, we take comments blocked under Rule 1 of the 24sata comment moderation rules to be spam (see Section~\ref{sec:dataset}).} we find that on average, spam comments are longer than other comments. We observe a similar pattern across different sections. %

Next, we measure lexical diversity using mean-segmental type-token ratio (MSTTR). The MSTTR is computed as the mean of type-token ratio for every 1000 tokens in a dataset to control for dataset size \cite{van2018varying}. From Table~\ref{tab:msttr}, we see that non-blocked comments have higher MSTTR (i.e.\ higher lexical diversity) than blocked comments (0.62 vs 0.46). %
However, when we again divide blocked comments into spam and non-spam, we observe that non-spam blocked comments have a similar MSTTR to non-blocked comments (0.61 vs 0.62), while spam comments have much lower MSTTR (0.35 vs 0.61). This suggests that blocked comments (excluding spam) have as rich a vocabulary as non-blocked. Again, we see a similar pattern across different news sections.

\begin{table}[h!]
\centering
\begin{tabular}{l r r } 
 \hline
 {} & \textbf{Mean length} & \textbf{MSTTR} \\
 \hline
 All & 23.06 & 0.61 \\ 
 \hline
 Non-blocked & 23.01 & 0.62 \\
 Blocked & 23.65 & 0.46 \\
 \hline
 Blocked (non-spam) & 19.16 & 0.61 \\
 Blocked (Spam only) & 28.23 & 0.35 \\  
\hline
\end{tabular}
\caption{Mean-segmental TTR and average length of comments}
\label{tab:msttr}
\end{table}

Now we look at the top bigrams of each class. We collect all bigrams that occur at least 50 times and rank them according to their pointwise mutual information (PMI) score. In general, we do not see many overlaps between the top bigrams of blocked and non-blocked comments across the different sections. Bigrams in blocked comments indicate spam messages such `iskustva potrebnog' (\textit{experience required}), `redoviti student' (\textit{full-time student}) and `prilika pružila' (\textit{opportunity given}). Removing spam comments, we encounter bigrams used for swearing such as
`pas mater' (\textit{damn it}) and `jedi govna' (\textit{eat sh*t}). %
In the non-blocked comments, the top bigrams are more relevant to the section they appear in. For instance, in the Vijesti section, top bigrams include `new york', `porezni obveznici' (\textit{taxpayers}) and `naftna polja' (\textit{oil fields}) while in Sports, top bigrams include `all star', `grand slam' and `man utd'. %

This suggests that the content of blocked comments tends to share commonalities across sections more than non-blocked comments; but again, these commonalities may be mostly within the spam category, with other blocked categories being more topic-dependent. Our next step therefore is to examine the use of topic modelling to capture these dependencies, with a view to using topic information to improve a moderation classifier. %

\section{Topic Modelling}
\label{sec:topic}

We now apply a topic model to gain insight into what characterises a blocked comment and a non-blocked one, and whether this varies between different sections where different subjects are discussed.

\subsection{Topic Model}
\label{sec:topicModel}
We use the Embedded Topic Model \citep[ETM,][]{dieng-etal-2020-topic} as our topic model since it has been shown to outperform regular LDA and and other neural topic modelling methods such as NVDM \citep{miao2016neural}. We also want to take advantage of ETM's ability to incorporate the information encoded in pretrained word embeddings trained on vast amounts of data to produce more coherent topics. In the ETM, the topic-term distribution for topic $k$, $\beta_{k}$, is induced by a matrix of word embeddings $\rho$  and its respective topic embedding $\alpha_{k}$ which is a point in the word embedding space:
\begin{equation}
\label{eq:topic-distribution}
\beta_{k} = softmax(\rho^{T}\alpha_{k}) 
\end{equation}
The topic embeddings are learned during topic inference while the word embeddings can be pretrained or also learned during topic inference. In this work, we use pretrained embeddings.

The document-topic distribution of a document $d$, $\theta_{d}$, is drawn from the logistic normal distribution whose mean and variance come from an inference network:
\begin{equation}
\label{eq:topic-theta}
\theta_{d} \sim LN(\mu_{d}, \sigma_{d}) 
\end{equation}

Given a trained ETM, we can infer the \textbf{document-topic distribution (DTD)} of an unseen document. In addition, we can also compute a \textbf{document-topic embedding (DTE)} as the weighted sum of the embeddings of the topics in a document, where the weight corresponds to the probability of the topic in that document:

\begin{equation}
\label{eq:topic-embedding}
DTE = \sum_{k=0}^{K} \alpha_{k} \theta_{d,k}
\end{equation}
where $\alpha_{k}$ is the topic embedding of topic $k$, and $\theta_{d,k}$ is the probability of topic $k$ in doc $d$.

\subsection{Topic Analysis}
\label{sec:analysis-topic}

Now we analyse the usage of topics in our test set. We trained the ETM for 100 topics on the training set and inferred the topic distributions of the comments in the test set. For analysis, we extract the top topics in a set of comments. To do this, we take the mean of the topic distributions over the comments in the set and rank the topics according to their weight in this mean distribution. We then take the top 15 topics for analysis because this is the average number of topics in a comment with a non-zero probability in our test set. Note that in this analysis we only use the document-topic distributions and not the document-topic embeddings. To more easily discuss the topics here we provide concise labels for each topic as interpreted by a native speaker. Automatic labelling of topics is a non-trivial task and an area of active research \cite{bhatia2016automatic, alokaili2020automatic, popa2021bart}.

\begin{figure}[!ht]
    \centering
    \includegraphics[width=0.45\textwidth]{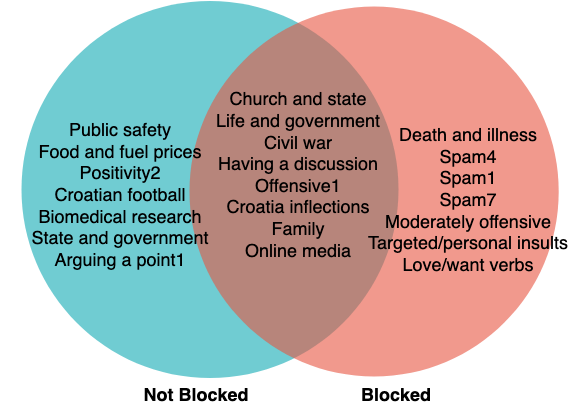}
    \caption{Top topics of the blocked and non-blocked comments for the entire test set.}
    \label{fig:venn-all}
\end{figure}

First, we examine the prevalent topics in the blocked and non-blocked comments, separately. The top topics of non-blocked comments cover a diverse range of subjects from politics to football while the top topics in blocked comments are dominated by spam and offensive language (Figure~\ref{fig:venn-all}). However, we also see many topics shared between blocked and non-blocked comments. \footnote{All 100 topics and labels are available at \url{https://github.com/ezosa/topic-aware-moderation}}.

Next we illustrate how different topics intersect and diverge between blocked and non-blocked comments across sections by looking at the top topics of two thematically-different sections, Lifestyle and Politika (\textit{Politics}). %

\begin{figure}[!ht]
    \centering
    \includegraphics[width=0.45\textwidth]{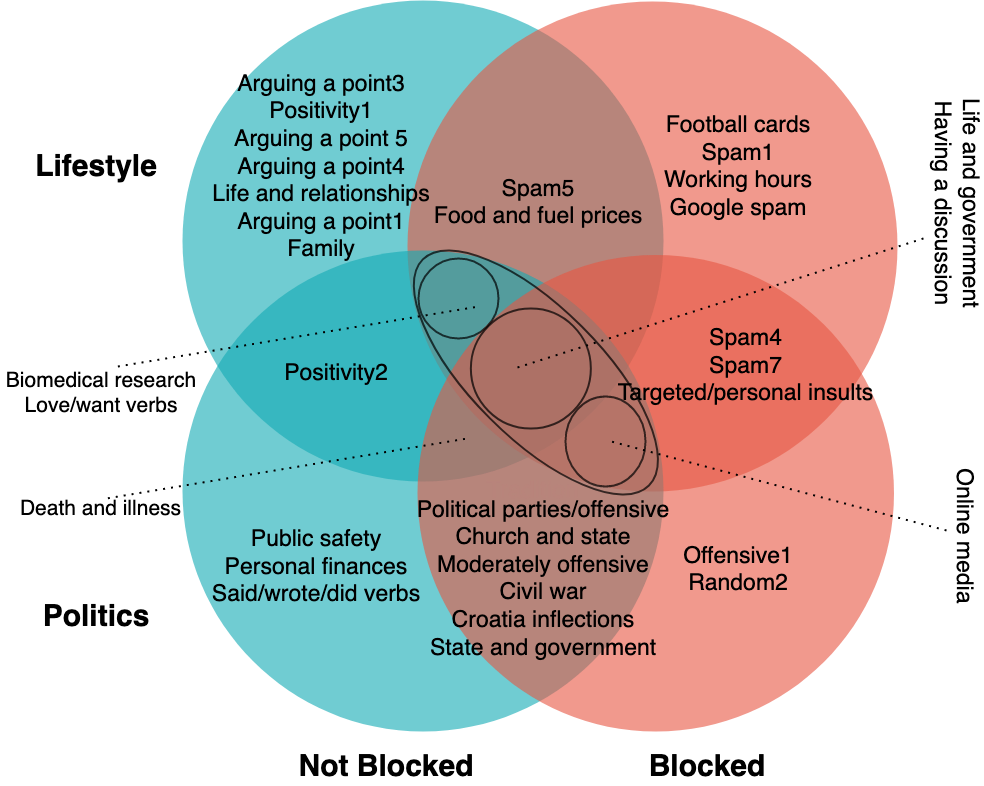}
    \caption{Top topics of the blocked and non-blocked comments in the Lifestyle and Politics sections.}
    \label{fig:venn-lifestyle-politics}
\end{figure}

Figure~\ref{fig:venn-lifestyle-politics} shows the top topics of these sections and the intersections between them. In Politics, blocked comments tend toward spam and targeted insults. Non-blocked topics include public safety and finances. However, we also see that more than half of the top topics overlap between blocked and non-blocked. This suggests that, thematically, there isn't a very clear distinction between blocked and non-blocked comments in the Politics section.

In Lifestyle, blocked topics are dominated by spam and while there are topics on offensive insults, they are not as prevalent as the spam-related ones. The non-blocked topics are about family and relationships and commenters arguing with each other. Compared to Politics, we see a clearer distinction between topics in blocked and non-blocked in this section. In terms of topic overlaps between Lifestyle and Politics, blocked comments in both sections are dedicated to spam and insults while non-blocked comments focus on positive sentiments. 

The combination of certain topics also provide an indication of the classification of the comment. For instance, we notice the use of topics about football cards in comments that do not do not discuss the sport (for instance, football cards as a topic is prominent in the blocked Lifestyle comments). It turns out that some commenters use the red and yellow cards from football as metaphors for being banned or having their comments blocked by moderators (12\% of comments that use these metaphors are blocked by moderators). On the other hand, comments that use the football cards topics \textit{and} any of the sports-related topics are likely to be a genuine discussion of football (only 5\% of such comments are blocked by moderators). We show some examples of these comments in Table~\ref{tab:sample-comments}. 

So clearly there is a distinction between the usage of topics in the non-blocked and blocked comments. We therefore think it is a good idea to propose a model which incorporates topic information into a comment moderation classifier.

\section{Topic-aware Classifier}
\label{sec:method}
\begin{figure*}
    \centering
    \includegraphics[width=0.9\textwidth]{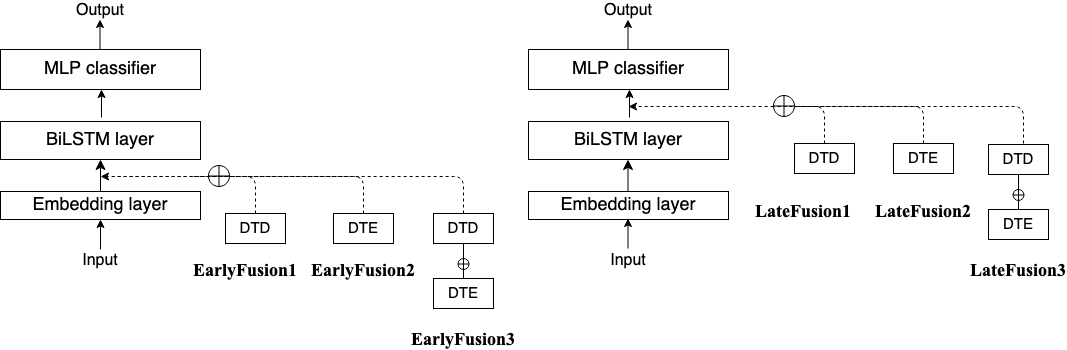}
    \caption{Architectures combining text and topic features. DTD is the topic distribution of a document while DTE is the topic embedding.}
    \label{fig:network_arch}
\end{figure*}

Our aim is to improve comment moderation predictions by combining textual features with document-level semantic information in the form of topics. To this end, we test several model architectures that combine a language model with topic features. %

For the comment text representation, we use a bidirectional LSTM~\citep[BiLSTM,][]{schuster1997bidirectional}. The comment text is given as input to an embedding layer then a BiLSTM layer where the output of the final hidden state is taken as the encoded representation of the comment.
For the topic representations, we use the topic distributions (DTD) and topic embeddings (DTE) discussed in Section~\ref{sec:topicModel}. 

We propose two fusion mechanisms to combine the text and topic representations: \textit{early} and \textit{late} fusion. In early fusion, topic features are concatenated with the output of the embedding layer and then passed to the BiLSTM layer. In \textbf{EarlyFusion1 (EF1)}, only DTD is concatenated with the word embeddings; \textbf{EarlyFusion2 (EF2)} uses DTE instead of DTD; and \textbf{EarlyFusion3 (EF3)} uses both DTE and DTD. In late fusion, topic features are concatenated with the output representation of the BiLSTM layer, and passed to the MLP for classification. Again, \textbf{LateFusion1 (LF1)} uses DTD;  \textbf{LateFusion2 (LF2)} uses DTE; and \textbf{LateFusion3 (LF3)} uses both. Figure~\ref{fig:network_arch} shows the architectures.

Our model is inspired by the Topic Compositional Neural Language Model~\citep[TCNLM,][]{wang2018topic} and the Neural Composite Language Model~\citep[NCLM,][]{chaudhary2020explainable} that incorporate latent document-topic distributions with language models. Both of these models simultaneously learn a topic model and a language model through a joint training approach. The NCLM introduced the use of word embeddings to generate an explanatory topic representation for a document in addition to the document-topic proportions. In our work, instead of using the word embeddings of the top words of the latent topics of a document (where the number of top words is a hyperparameter), we leverage the topic embeddings learned by ETM and combine them with the document-topic proportions to produce the document-topic embeddings (DTE). Also unlike the TCNLM and NCLM, we use pre-trained topics in our model so as to easily de-couple and analyse the influence of topics in the classifier performance. Another related work is TopicRNN~\cite{dieng2016topicrnn}, a model that uses topic proportions to re-score the words generated by the language model. The topics generated by this model, however, have been shown to have lower coherences compared to NCLM \cite{chaudhary2020explainable}. 

\section{Experimental Setup}
\label{sec:exp}

\paragraph{Dataset} As discussed in Section~\ref{sec:data-selection}, we use the 2018 data as the training and validation sets of our topic-aware classifier and the 2019 data as the test set. Details of the train and validation sets are shown in Table~\ref{tab:dataset-train} and the test set in Table~\ref{tab:dataset-2019}.

\paragraph{Baseline models}
To assess how topic information improves comment classification, we use as baselines the following models trained only on text \textit{or} topics:
\begin{compact_itemize}
    \item \textbf{Text only}: a classifier with BiLSTM \& MLP layers, similar to Figure~\ref{fig:network_arch} but with comment text alone as input.
    \item \textbf{Document-topic distribution (DTD)}: MLP only, document-topic distributions as input.
    \item \textbf{Document-topic embedding (DTE)}: MLP only, document-topic embeddings as input.
    \item \textbf{DTD+E}: MLP only, concatenated document-topic distributions and embeddings.
\end{compact_itemize}

\paragraph{Hyperparameters} 
We use 300D word2vec embeddings, pretrained on the Croatian Web Corpus~\citep[HrWAC,][]{ljubevsic2011hrwac,snajder-2014-derivbase}, for training the ETM and to initialize the embedding layer of the BiLSTM. The ETM is trained for 500 epochs for 100 topics using the default hyperparameters from the original implementation \footnote{\url{https://github.com/adjidieng/ETM}}. The BiLSTM is composed of one hidden layer of size 128 with dropout set to 0.5. The MLP classifier is composed of one fully-connected layer, one hidden layer of size 64, a ReLU activation, and a sigmoid for classification with the classification threshold set to $0.5$. We use Adam optimizer with $lr = 0.005$. We train all classifiers for 20 epochs with early stopping based on the validation loss.

\section{Results}
\label{sec:results}

In Table~\ref{tab:results}, we present the performance of the baselines and proposed models, measured as macro F1-scores. All models that combine text and topic representations perform better than the models that use only text \textit{or} topics. Of the baseline models, the DTD model performs comparatively better than the DTE and DTD+E models, and surprisingly performs almost as well as the Text-only model; however, we show in Section~\ref{sec:conf} below that DTD is much less confident in its predictions than the Text-only model. Overall, the best performing model is LF1, which improves the Text-only model's performance by $+4.4\%$ (67.37\% vs 62.97\%); and improves by a similar amount over \citeauthor{shekhar-etal-2020-automating}'s results using mBERT (macro-F1 score 62.07 for year 2019).

Interestingly, we see a wide variation in performance across news sections. We observe that comments in Lifestyle and Tech are the easiest to classify (best F1 over 72.00) while Politika (\emph{Politics}) is the most difficult (best F1 around 61.61). The main cause appears to be that Lifestyle and Tech have the highest proportion of spam comments: on average, 49.44\% of blocked comments in the test set are spam, but for Lifestyle and Tech this number rises to 77.25\% and 69.63\%, respectively. As for the Politics section, the most likely reason the comments are difficult to classify is that, excluding spam, there is a high degree of overlap in the subjects discussed in the blocked and non-blocked comments (see the topic analysis in Section~\ref{sec:analysis-topic}). %

\begin{table*}[t]
\centering
\resizebox{\textwidth}{!}{
\begin{tabular}{ l | c | c | c | c || c | c | c | c | c | c }
 {\textbf{Section}} & {\textbf{Text}} & \multicolumn{3}{c ||}{\textbf{Topics only}} & \multicolumn{6}{c }{\textbf{ Text+Topic Combinations}} \\
 {\textbf{$~-$ Subsection}} & \textbf{only} & \textbf{DTD} & \textbf{DTE} & \textbf{DTD+E} & \textbf{EF1} & \textbf{EF2} & \textbf{EF3}  & \textbf{LF1} & \textbf{LF2} & \textbf{LF3}\\
 \hline 
 
All	&   62.97	& 62.20	& 59.3	& 58.33	& 66.33	& 66.58	& 65.61	& \textbf{67.37}	& 66.22	& 66.95 \\
\hline \hline
Kolumne &  59.86	& 59.65	& 56.25	& 55.33	& 62.40	& 62.90	& 63.13	& 63.25	& 62.38	& \textbf{63.6}\\
Lifestyle & 69.21	& 70.07	& 65.93	& 64.47	& 72.73	& 70.9	& 69.36	& 72.00	& 72.39	& \textbf{72.92}\\
Show      & 61.97	& 61.30	& 58.62	& 57.60	& 65.24	& 65.63	& 64.26	& \textbf{66.50}	& 65.00	& 65.86\\
Sport     & 63.22	& 61.42	& 58.61	& 57.90	& 67.11	& 67.86	& 66.74	& \textbf{68.26}	& 67.14	& 67.82\\
Tech	    & 64.87	& 66.37	& 63.17	& 62.55	& 67.72	& 68.74	& 67.65	& 68.76	& 67.68	& \textbf{69.15}  \\ 
\hline %
Vijesti (\textit{News}) & 62.38	& 61.49	& 58.79	& 57.77	& 65.58	& 65.99	& 65.24	& \textbf{66.77}	& 65.53	& 66.24  \\
\hline
$~-$ Crna kronika &  64.67	& 63.98	& 61.03	& 59.84	& 68.10	& 68.88	& 68.11	& \textbf{69.60}	& 67.89	& 68.88\\
$~-$ Hrvatska	   &  63.61	& 63.50	& 60.10	& 58.93	& 67.24	& 66.86	& 65.95	& 67.90	& 67.12	& \textbf{67.95} \\
$~-$ Politika	   &  57.93	& 56.49	& 54.95	& 54.20	& 60.51	& 61.52	& 60.84	& \textbf{61.61}	& 60.63	& 61.30\\
$~-$ Svijet       &  63.58	& 62.55	& 59.62	& 58.35	& 66.83	& 66.95	& 66.33	& 68.44	& 67.21	& 67.57\\

\hline
\end{tabular}
}
\caption{Classifier performance measured as macro-F1.}
\label{tab:results}
\vspace*{-2mm}
\end{table*}

\subsection{Analysis of Classifier Outputs}
\label{sec:analysis-outputs}

\begin{table*}[!ht]
\resizebox{\textwidth}{!}{
{\normalsize
\begin{tabular}{p{9cm}|p{1cm}|p{1.6cm}|p{1.5cm}|p{3cm}}
\hline
\textbf{Comment} & \textbf{Label} &\textbf{Text-only} & \textbf{LF1} & \textbf{Top topics} \\ 
\hline

1.  konačno. gamad lopovska crno bijela prevarantska (\emph{finally. the black and white cheating thieving bastards}) &    	1   &	1 (0.501)     &	1 (0.687) & Arguing a point, Political parties (offensive)\\
\hline

2.  ...dobro jutro,moze crveni karton za novinara koji je osmislio naslov ;-) (\emph{... good morning, how about a red card for the journalist who came up with this title ;-)}) &	1	&   0 (0.315) &   0 (0.456) & Football cards\\
\hline

3.  Ne bum komentiral, dosta mi je kazni od žutih i crvenih kartona. Strah me je cenzure i bradate cure. (\emph{No comment, I'm tired of getting yellow and red cards. I'm afraid of censorship and bearded ladies.}) & 0 & 0 (0.054) & 0 (0.335) & Football cards, Random\\
\hline

4.  Koji kurac Rumunjski sudac ne da koji karton više Čehima. Pa svake tri minute sa leđa sruše Olma !!!! \emph{(Why the fuck does the Romanian referee not give a few cards more to the Czechs, They tackle Olm from behind every three minutes.)} & 0 & 0 (0.303) & 1 (0.587) & Targeted/personal insults \\
\hline

5.  Baš ste jadnici kao i ovi sa 24sata koji u ovome uživaju ! (\emph{All of you are lame  as well as those from 24sata who enjoy this.})  &  1   &	0 (0.171) &   0 (0.229) & Online media, Moderately offensive \\
\hline

6.  Google sada plaća između 15.000 i 30.000 dolara mjesečno za rad na mreži od kuće. Pridružio sam se ovom poslu prije 3 mjeseca i zaradio 24857 dolara u prvom mjesecu ovog posla. %
$>>>$ URL (\emph{Google now pays between 15.000 and 30.000 dollars per month for working remotely from home. I started this job 3 months ago and made 24857 dollars in the first month of this job. %
$>>>$ URL})  & 0 & 1 (0.67) & 1 (0.90) & Spam4\\
\hline

\end{tabular}
}
}
\caption{Sample comments and classifier decisions.}
\label{tab:sample-comments}
\end{table*}

In general, we observe that blocked comments tend to use similar topics across different sections while non-blocked comments have more diverse topics. Of the nine sections that we analyzed, there are five topics that are prominent in blocked comments in all sections (`Targeted/personal insults', `Spam4', `Spam7', `Online media', and, `Having a discussion') and only three topics prominent in non-blocked comments (`Having a discussion', `Online media', and, `Life and government'). This suggests that blocked comments are more semantically-coherent across sections than non-blocked ones. In contrast, topics in non-blocked comments tend to be more relevant to their respective sections: for instance, family and relationships are not discussed a lot in the Politics section, while Lifestyle commenters do not tend to talk about political issues. 

The higher topical coherence then of blocked comments explains why a text classification approach can achieve reasonable performance; but the variation in blocked comment content between some sections explains why adding topic information improves our classification results.

\begin{figure}[!ht]
    \centering
    \vspace{-5mm}
    \includegraphics[width=0.5\textwidth]{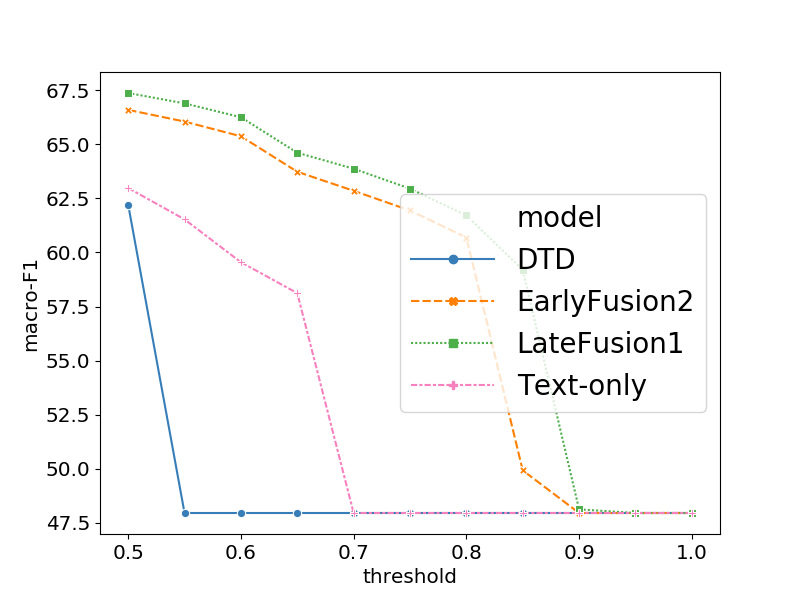}
    \caption{Confidence of the top performing models.}
    \label{fig:classifier-confidence}
    \vspace{-2mm}
\end{figure}

Next, we analyze the confidence of classifiers and examine some of the outputs of the models. To analyze confidence, we gradually increase the classification threshold from 0.5 to 1.0 in increments of 0.05. For every new threshold, we plot the macro-F1 for the different models (Figure~\ref{fig:classifier-confidence}). We compare the confidence of four models: DTD, Text-only, EF2 (the strongest early fusion model), and LF1 (the overall best-performing model). We find that the most confident model is LF1 and the least confident is DTD. The two fusion classifiers display similar levels of confidence. The Text-only classifier is not as confident as the fusion classifiers but still more confident than DTD. This suggests that adding topic features to text not only improves performance, it also increases classifier confidence. 

In Table~\ref{tab:sample-comments} we give some examples of comments and the classifier decisions of the Text-only classifier and LF1 (our best-performing fusion model) and their top topics (topics with $prob > 0.10$). The first example contains swearing which both models pick up on and classify as blocked although LF1 is more confident in its decision then Text-only. In the second example, both models predict the wrong label but LF1 treats this as a borderline case because it is targeted at the moderators. However since this is only a mild provocation of the moderators, this might be a case where the gold label is incorrect. The topics also pick up on the fact that this comment talks about football cards but only has a tenuous connection to the sport (``getting a red card'' is an expression used for ``being banned''). In contrast, the third comment also uses the banning sense of ``card'' but is not directed at anyone, and is thus labeled as 0 (non-blocked), which both models get right. Again the topics indicate that the comment is not really about the sport. The fourth example shows a case where ``cards'' are mentioned in their standard football sense but also contains a swear word, making the gold label of 0 (non-blocked) questionable. The better performance of LF1 on such examples, compared to Text-only, implies that LF1 is better aware of the different semantics of ``card'' (sports-related  vs. metaphorical), likely due to added topic information. 

The fifth example contains a moderately offensive insult that is not directed at any single group except the 24sata readership in general. One reason why both classifiers do not get this right is that the word \textit{jadnici} is not strong enough to be considered offensive. Finally the last example is clearly a spam comment  that both classifiers correctly classify but for which the gold label is incorrect. 

Overall, compared to the Text-only model, we find that LF1 more often than not improves the confidences (and sometimes the classification), especially in cases in which the gold label is clear. This is valuable in practice, as better confidences might lead to better prioritisation of comments for manual moderation, reducing the time required to remove the most problematic ones.

\section{Conclusion}
\label{sec:conf}
In this work, we propose a model to combine document-level semantics in the form of topics with text for comment moderation. Our analysis shows that blocked and non-blocked comments have different linguistic and thematic features, and that topics and language use vary considerably across news sections, including some variation in the comments that should be blocked. We also found that blocked comments tend to be more semantically coherent across sections than non-blocked ones. We therefore see that the use of topics in our model improves performance, and gives more confident outputs, over a model that only uses the comment text. The model also provides topic distributions, interpretable as keywords, as a form of an explanation of its prediction. As future work, we plan to incorporate comment, article, and user metadata into the model.

\section*{Acknowledgements}
This work has been supported by the European Union Horizon 2020 research and innovation programme under grants 770299 (NewsEye) and 825153 (EMBEDDIA), and by the UK EPSRC under grant EP/S033564/1.

\section*{Ethics and Impact Statement}
\label{sec:impact-statement}
\paragraph{Data} The dataset and annotations are provided by the publisher of 24sata.hr, Styria Media Group, for research purposes and deposited in the CLARIN repository. The authors of the comments are anonymised. The researchers used the data as-is and did not modify or add annotations. 

\paragraph{Intended Use} The models we present here are intended to assist comment moderators in their work. We do not recommend that the model be deployed in the moderation process without a human-in-the-loop.

\paragraph{Potential Misuse} The models and the analysis of their performance we provide in this paper could be used by malicious actors to gain an insight into the comment moderation process and find loopholes in the process. However, we think such a risk is unlikely and the impact it might have outweighs the potential benefits of models intended to assist human moderators such as the ones we present here. 

\bibliographystyle{acl_natbib}
\bibliography{reference}

\end{document}